\documentclass[journal]{IEEEtran}
\usepackage{amsmath,amsfonts}

\usepackage{algorithm}
\usepackage{algorithmicx}
\usepackage{algpseudocode}

\usepackage{array}
\usepackage[caption=false,font=normalsize,labelfont=sf,textfont=sf]{subfig}
\usepackage{textcomp}
\usepackage{stfloats}
\usepackage{url}
\usepackage{verbatim}
\usepackage{graphicx}
\usepackage{cite}
\usepackage{booktabs}
\usepackage[table]{xcolor} 
\usepackage{multirow}
\usepackage{hyperref}
\usepackage{amsmath,amsfonts,amssymb,bm,mathtools}

\def\ie{\emph{i.e.}}

\def\1{\bm{1}}

\def\vepsilon{{\bm{\epsilon}}}

\def\vf{{\bm{f}}}
\def\vg{{\bm{g}}}

\def\vz{{\bm{z}}}

\def\mC{{\bm{C}}}

\def\mI{{\bm{I}}}

\def\mK{{\bm{K}}}

\def\mM{{\bm{M}}}

\def\mO{{\bm{O}}}

\def\mQ{{\bm{Q}}}

\def\mV{{\bm{V}}}

\DeclareMathAlphabet{\mathsfit}{\encodingdefault}{\sfdefault}{m}{sl}
\SetMathAlphabet{\mathsfit}{bold}{\encodingdefault}{\sfdefault}{bx}{n}

\def\gL{{\mathcal{L}}}

\newcommand{\E}{\mathbb{E}}

\newcommand{\softmax}{\mathrm{softmax}}

\newcommand{\reff}[1]{Fig.~\ref{fig:#1}}
\newcommand{\reft}[1]{Table~\ref{table:#1}}
\newcommand{\refe}[1]{Eq.~\ref{eqn:#1}}
\newcommand{\refs}[1]{Sec.~\ref{sec:#1}}
\newcommand{\refa}[1]{Algorithm~\ref{alg:#1}}

\newcommand{\lsds}{\gL_{\mbox{\tiny SDS}}}
\newcommand{\std}[1]{\sqrt{1 - \alpha_{#1}}}

\newcommand{\scale}[1]{\sqrt{\alpha_{#1}}}

\newcommand{\zpred}[1]{\hat{\vz}_0^{(#1)}}

\newcommand{\zpi}{\vz_{\pi}}

\newcommand{\ysrc}{y^{\text{src}}}
\newcommand{\ytgt}{y^{\text{tgt}}}

\newcommand{\Qsrc}{\mQ_{\text{src}}}

\newcommand{\Qtgt}{\mQ_{\text{tgt}}}
\newcommand{\Ktgt}{\mK_{\text{tgt}}}
\newcommand{\Vtgt}{\mV_{\text{tgt}}}

\newcommand{\thetasrc}{\theta^{\text{src}}}
\newcommand{\thetat}[1]{\theta^{(#1)}}

\newcommand{\zsrc}[1]{\vz^{\text{src}}_{#1}}
\newcommand{\ztgt}[1]{\vz^{\text{tgt}}_{#1}}

\newcommand{\zsrccons}[1]{\hat{\vz}^{\text{src}}_{#1}}
\newcommand{\ztgtcons}[1]{\hat{\vz}^{\text{tgt}}_{#1}}

\newcommand{\epsilonsrc}{\vepsilon^{\text{src}}_{\phi}}
\newcommand{\epsilontgt}{\vepsilon^{\text{tgt}}_{\phi}}

\def\vdelta{{\bm{\delta}}}

\begin{document}

\title{TrAME: Trajectory-Anchored Multi-View Editing for Text-Guided 3D Gaussian Manipulation}

\author{Chaofan Luo, Donglin Di, Xun Yang$^\star$, Yongjia Ma, Zhou Xue, Wei Chen, Xiaofei Gou, \\
Yebin Liu$^\star$,~\IEEEmembership{Member,~IEEE}


\thanks{$^\star$Corresponding authors: Xun Yang and Yebin Liu.}

\thanks{Chaofan Luo and Xun Yang are with School of Information Science and Technology, University of Science and Technology of China, Hefei, 230026, Anhui, China. Email: \href{mailto: luocfprime@gmail.com}{luocfprime@gmail.com}, \href{mailto: xyang21@ustc.edu.cn}{xyang21@ustc.edu.cn}
}


\thanks{Donglin Di, Yongjia Ma, Chaofan Luo, Zhou Xue, Wei Chen, and Xiaofei Gou are with Space AI, Li Auto, 101399, Beijing, China.
Email: \href{mailto:didonglin@lixiang.com, mayongjia@lixiang.com, xuezhou08@gmail.com, chenwei10@lixiang.com, gouxiaofei@lixiang.com}{\{didonglin, mayongjia, chenwei10, gouxiaofei\}@lixiang.com, luocfprime@gmail.com, xuezhou08@gmail.com}
}


\thanks{Yebin Liu is with Department of Automation,
Tsinghua University, Beijing 100084, China. Email: \href{mailto: liuyebin@mail.tsinghua.edu.cn}{liuyebin@mail.tsinghua.edu.cn}
}

}



\maketitle

\begin{abstract}

Despite significant strides in the field of 3D scene editing, current methods encounter substantial challenge, particularly in preserving 3D consistency in multi-view editing process.
To tackle this challenge, we propose a progressive 3D editing strategy that ensures multi-view consistency via a Trajectory-Anchored Scheme (TAS) with a dual-branch editing mechanism.
Specifically, TAS facilitates a tightly coupled iterative process between 2D view editing and 3D updating, preventing error accumulation yielded from text-to-image process. Additionally, we explore the relationship between optimization-based methods and reconstruction-based methods, offering a unified perspective for selecting superior design choice, supporting the rationale behind the designed TAS.
We further present a tuning-free View-Consistent Attention Control (VCAC) module that leverages cross-view semantic and geometric reference from the source branch to yield aligned views from the target branch during the editing of 2D views.
To validate the effectiveness of our method, we analyze 2D examples to demonstrate the improved consistency with the VCAC module.
Further extensive quantitative and qualitative results in text-guided 3D scene editing indicate that our method achieves superior editing quality compared to state-of-the-art methods.
We will make the complete codebase publicly available following the conclusion of the review process.

\end{abstract}

\vspace{-5pt}

\section{Introduction}

\label{sec:introduction}

\vspace{-5pt}

3D technologies are pivotal in various fields, such as virtual reality (VR), the film and gaming industries, and 3D design \cite{zhu2024hifa,he2023customize,mirzaei2024reffusion,feng2024lsk3dnet,feng2024lsk3dnet,feng2023clustering}.
Recent advances in Neural Radiance Fields (NeRF) and 3D Gaussian Splatting (3DGS) have pushed the boundary of 3D representation, enabling versatile 3D applications \cite{mildenhall2021nerf,muller2022instant,NEURIPS2021_e41e164f,dellaert2020neural,kerbl20233d,ye2023gaussian,wu2024recent,chen2024survey,meng2021towards}.
Leveraging the powerful generative capabilities of text-to-image diffusion models \cite{brooks2023instructpixpix, rombach2022high}, text-guided 3D editing now allows for intricate adjustments in shape, style, texture, and lighting, marking a significant advancement in the flexibility of 3D scene manipulation \cite{Haque_2023_ICCV,kamata2023instruct,he2023customize,mirzaei2024reffusion,khalid2023latenteditor,chen2024dge,khalid20243dego,michel2024object,huang2023customize}.

One key challenge in 3D editing is maintaining multi-view consistency when applied to real scenes \cite{song2023efficient,wu2024gaussctrl}, which is essential for preventing visual artifacts and inconsistent appearances when observed from different view angles.
Current studies of 3D editing can be divided into two categories, namely optimization-based and reconstruction-based methods.
The optimization-based 3D editing methods \cite{koo2023posterior,sella2023vox,li2024focaldreamer,10.1145/3610548.3618190,kamata2023instruct} usually adopt the Score Distillation Sampling (SDS) loss \cite{poole2023dreamfusion} or its modified versions \cite{hertz2023delta, park2024ednerf}. 
They often suffer from sub-optimal editing quality such as over-saturation, over-smoothing, and a lack of diversity due to the misalignment between the randomly sampled timestep $t$ in the SDS optimization process and the corresponding timestep $t$ in the diffusion sampling process \cite{huang2023dreamtime,zhu2024hifa}. 
The second category of methods, reconstruction-based 3D editing methods \cite{Haque_2023_ICCV,chen2023gaussianeditor,wu2024gaussctrl,song2023efficient,khalid2023latenteditor}, employ readily available 2D editing diffusion models to edit 3D scene.
However, 2D diffusion models face challenges in achieving multi-view consistency due to their intrinsic limitation of independently processing each view.

In this paper, to mitigate the above-mentioned issues, we first theoretically unveil the relation between these two branches of methods, demonstrating that optimization-based methods are special cases of reconstruction-based methods.
Specifically, we uncover an intrinsic equivalence between the \textit{optimization process} of SDS \cite{poole2023dreamfusion} and an \textit{iterative reconstruction process} that aims to match pseudo-ground-truths obtained from the Denoising Diffusion Consistent Model (DDCM) \cite{xu2024inversionfree}.
This analysis unites these two branches of 3D editing methods, offering a unified perspective for making superior design choices.
Based on the analysis, we introduce a progressive 3D editing framework named ``Trajectory-Anchored Multi-View Editing (TrAME)''.
The core idea of this framework is the Trajectory-Anchored Scheme (TAS), a unified paradigm designed to gradually consolidate incremental 2D view edits onto 3D scenes.
Inspired by an empirical practice of utilizing feedback from rendered images of the updated 3DGS to correct errors that arise during the 2D editing process\cite{chen2024generic,zuo2024videomv}, we design TAS to enhance the interplay between the image editing process and the 3DGS update process, allowing for active rectification of inconsistencies within 2D multi-view edits through the 3D constrained rendering process (\ie, rendering in the loop).
Utilizing a modified DDCM editing trajectory \cite{xu2024inversionfree}, we incrementally generate pseudo-ground-truths for progressive 3D Gaussian editing.
Furthermore, to enhance 3D-consistency during the 2D view editing process, we design a VCAC module with a dual-branch editing strategy.
The edited views maintain \textit{structural multi-view consistency} via self-attention queries injected from the source branch into the target branch during the early stages of diffusion.  
To enhance \textit{semantic multi-view consistency}, we facilitate self-attention Key-Value (KV) propagation and KV reference mechanisms, propagating the self-attention keys and values of keyframes within the same context and inflating the self-attention keys and values across different keyframes for a mutual reference.

We conduct extensive experiments to demonstrate the effectiveness of our proposed TAS strategy and VCAC module.
Both qualitative and quantitative results of 3D editing highlight the superiority of our approach in comparison to state-of-the-art methods.
Additionally, the qualitative results from ablation studies demonstrate enhanced consistency with our proposed TAS and VCAC module.
Our main contributions are summarized as follows:

$\bullet$
Our theoretical analysis bridges the gap between optimization-based and reconstruction-based editing methods, offering a unified perspective for selecting superior design choices.

$\bullet$
We propose a progressive 3D editing strategy that integrates a Trajectory-Anchored Scheme (TAS).
This design ensures multi-view consistency by tightly coupling the iterative processes of 2D view editing and 3D scene updating, thereby preventing error accumulation during text-to-image process.

$\bullet$
We present a tuning-free VCAC module that enhances the 3D consistency of editing results. 
The VCAC leverages cross-view semantic and geometric references from the source branch to yield aligned views in the target branch via query injection, KV propagation and KV reference mechanisms.

$\bullet$
Extensive experimental results in text-guided 3D scene editing demonstrate that our method achieves improved multi-view consistency editing results compared with state-of-the-art methods.
 

\section{Related Work}


In this section, we provide a concise overview of two primary branches of methodologies employed in the task of 3D scene editing: optimization-based methods and reconstruction-based methods.

Regarding \textbf{optimization-based 3D editing}, significant advancements have been made, building upon the foundational SDS loss introduced by DreamFusion \cite{poole2023dreamfusion}.
This has catalyzed the development of various innovative methods \cite{koo2023posterior,sella2023vox,li2024focaldreamer,10.1145/3610548.3618190,kamata2023instruct,park2024ednerf} aimed at refining 3D model editing via optimizing the SDS loss.
Notably, Vox-E \cite{sella2023vox} and DreamEditor \cite{10.1145/3610548.3618190} have harnessed explicit 3D representations, including voxels and meshes.
These approaches utilize cross-attention mechanisms to facilitate precise manipulations within specific regions of 3D models.
Furthermore, RePaint-NeRF \cite{zhou2023repaint} has advanced the application of SDS in 3D editing by integrating a semantic mask to guide and constrain modifications within the background elements.
In a similar vein, ED-NeRF \cite{park2024ednerf} has introduced an enhanced loss function specifically designed for 3D editing tasks. 
It extends the Delta Denoising Score (DDS) loss \cite{hertz2023delta} into the three-dimensional space, offering a more refined approach to editing.
Posterior distillation sampling \cite{koo2023posterior} aligns the identities of sampled targets with their corresponding sources by matching the stochastic latent variables obtained through DDPM inversion.
Despite these advancements, a common challenge still persists among methods that utilize randomly sampled timesteps $t$ within the SDS framework:
This approach tends to lead to deviations from the intended diffusion sampling trajectories, which can result in compromised quality of the 3D edits.

For \textbf{reconstruction-based 3D editing}, InstructN2N \cite{Haque_2023_ICCV} alongside subsequent studies \cite{chen2023gaussianeditor,wu2024gaussctrl,song2023efficient,khalid2023latenteditor,chen2024consistdreamer}, have leveraged advancements in 2D diffusion editing techniques, notably InstructP2P \cite{brooks2023instructpixpix} and ControlNet \cite{zhang2023adding}, for enhancing scene updates through Iterative Dataset Update \cite{Haque_2023_ICCV}.
GaussianEditor \cite{chen2023gaussianeditor} and GaussCtrl \cite{wu2024gaussctrl} introduce methods for precise control in partial 3D editing.
GaussianEditor utilizes semantic tracing to segment the 3D Gaussians to be edited. 
GaussCtrl proposes depth-guided editing and attention-based latent alignment.
Despite their innovations, these methods often neglect inter-view correlations, resulting in a gap in maintaining semantic and structural consistency across multiple views, ultimately leading to error accumulation.
ViCA-NeRF \cite{dong2023vicanerf} adopts a depth-guided blending strategy to enforce 3D consistency across edited multi-views. However, this approach is prone to accumulating errors from depth estimation, frequently resulting in blurry and corrupted edits.
Our work bridges this gap by integrating Trajectory-Anchored Scheme to minimize error accumulation and introducing attention manipulation mechanisms to enforce structural and semantic consistency across views in 3D editing through our VCAC module.

\vspace{-8pt}

\section{Preliminaries}

\vspace{-5pt}

In this section, we first briefly review Score Distillation Sampling (SDS) and Denoising Diffusion Consistent Model (DDCM) before delving into the analysis of the relation between optimization-based editing and reconstruction-based editing.

\vspace{-5pt}

\subsection{Score Distillation Sampling}

\vspace{-5pt}

Score Distillation Sampling (SDS) \cite{poole2023dreamfusion}, also known as Score Jacobian Chaining (SJC) \cite{wang2023score}, is a commonly adopted technique in 3D asset generation and editing. It performs sampling in parameter space by optimizing a diffusion distillation loss function. Given a camera pose $c$ and the parameters of a 3D representation $\theta$, noisy latents $\vz_t$ are obtained by first rendering different views and then adding noise to the latents of the rendered images. The gradient with respect to the parameter of the 3D representation $\theta$ is defined as in \refe{sds_grad}, using a pretrained diffusion model $\vepsilon_\phi$ to predict the noise estimate conditioned by $y$ and match it against the ground truth noise $\vepsilon$. This gradient is backpropagated through a differentiable render function $\vg(\cdot, c)$, scaled by a weighting schedule $\omega(t)$.
\begin{equation}
\label{eqn:sds_grad}
    \nabla_{\theta} \lsds(\theta) \coloneq \E_{t,\vepsilon,c}\,\left[\omega(t) (\vepsilon_\phi(\vz_t,t,y) - \vepsilon) \frac{\partial\vg(\theta,c)}{\partial\theta}\right].
\end{equation}
Additionally, variant loss functions, such as Delta Denoising Score, are often used as a substitute for the vanilla SDS loss.

\vspace{-5pt}

\subsection{Denoising Diffusion Consistent Model}

\vspace{-5pt}

Denoising Diffusion Implicit Models (DDIMs) \cite{song2021denoising} characterize a family of generative models with non-Markovian forward processes.
Given a noisy latent $\vz_t$ with condition $y$ at timestep $t$, the sampling procedure of DDIM is formulated as:
\begin{equation}
    \begin{aligned}
        \vz_{t-1} =& \sqrt{\alpha_{t-1}} \underbrace{\left(\frac{\vz_t - \sqrt{1 - \alpha_t} \vepsilon_\phi(\vz_t,t,y)}{\sqrt{\alpha_t}}\right)}_{\text{ predicted } \zpred{t}} 
        \\
        &+ \underbrace{\sqrt{1 - \alpha_{t-1} - \sigma_{t}^2} \cdot \vepsilon_\phi(\vz_t,t,y)}_{\text{direction pointing to } \vz_t } + \underbrace{\sigma_{t} \vepsilon_t}_{\text{random noise}} ,
    \end{aligned}
\label{eqn:sample-ddim}
\end{equation}

where $\vepsilon_\phi$ is the noise predictor parameterized by parameter $\phi$ and $\alpha_t$ corresponds to diffusion schedule. Different instances of $\sigma_t$ in \refe{sample-ddim} yield different generative processes \cite{song2021denoising}.
Consider a special generative process where $\sigma_{t} \coloneq \sqrt{1 - \alpha_{t-1}}$.
The second term on the right-hand side of \refe{sample-ddim} is canceled out, leaving only $\zpred{t}$ and the newly added noise $\sigma_{t} \vepsilon_t$: 
\begin{equation}
        \vz_{t-1} = \sqrt{\alpha_{t-1}} \left(\frac{\vz_t - \sqrt{1 - \alpha_t} \vepsilon_\phi(\vz_t,t,y)}{\sqrt{\alpha_t}}\right) + \sqrt{1 - \alpha_{t-1}} \vepsilon_t \; ,
    \label{eqn:sample-ddcm}
\end{equation}
which is formulated as Denoising Diffusion Consistent Model (DDCM) \cite{xu2024inversionfree}.

\IEEEpubidadjcol
\vspace{-5pt}

\section{Methodology}
\label{sec:methodology}

\vspace{-5pt}

\begin{figure*}[tb]
\centering
  \includegraphics[width=\textwidth]{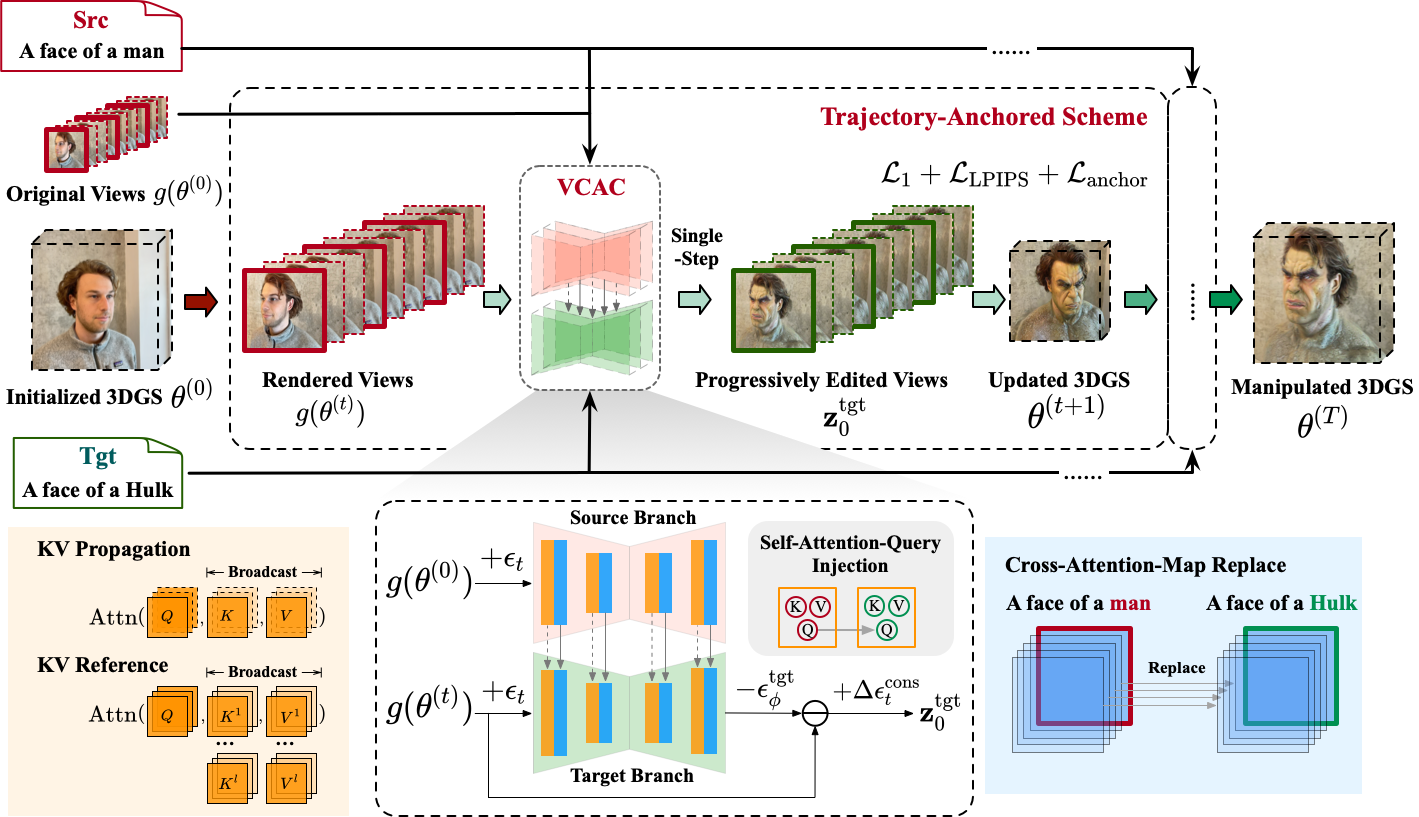}
  \vspace{-8pt}
  \caption{Illustration of the proposed method, Trajectory-Anchored Multi-View Editing for 3D Gaussian Splatting Manipulation (TrAME).
    Our method comprises a Trajectory Anchored Scheme (TAS) as well as a View-Consistent Attention Control (VCAC) module.
    Given a source prompt, a target prompt and the original 3DGS $\theta^{(0)}$ as input, the VCAC module can yield 3D-consistent and progressively edited views with a \textit{single-step inference} to update 3DGS. Conversely, the views rendered from the updated 3DGS correct minor inconsistencies from previous view edits and serve as inputs for subsequent steps, thereby preventing error accumulation from the 2D editing process. This process alternatively update the 2D views and 3DGS in a synchronized and progressive manner, producing the final edited 3DGS $\theta^{(T)}$.}
    \vspace{-12pt}
  \label{fig:framework}
\end{figure*}

We first analyze the relation between optimization-based and reconstruction-based 3D scene editing methods in detail.
Then, we elaborate on the proposed progressive Trajectory-Anchored Scheme (TAS) and the design of View-Consistent Attention Control (VCAC). 
The overall framework is illustrated in \reff{framework}.

\vspace{-15pt}

\subsection{Analysis on Optimization-based Editing}
\label{sec:analysis}

\vspace{-3pt}

We analyze the optimization process of score distillation sampling, and generalize it to its variants.
The gradient of the SDS loss function, as defined in \refe{sds_grad}, is equivalent to the gradient obtained by optimizing the rendered views to match a pseudo-ground-truth $\zpred{t}$:
\begin{equation}
    \begin{aligned}
    \nabla_{\theta} \lsds(\theta) &\coloneq \E_{t,\vepsilon,c}\,\left[\omega(t) \left(\vepsilon_\phi(\vz_t,t,y) - \vepsilon\right) \frac{\partial\vg(\theta,c)}{\partial\theta}\right]
\\
&= \E_{t,\vepsilon,c}\,\left[\omega(t) \frac{\scale{t}}{\std{t}}\left(\vz_\pi - \zpred{t}\right) \frac{\partial\vg(\theta,c)}{\partial\theta}\right] \label{eqn:pseudo-gt-1} \; ,
\end{aligned}
\end{equation}
where $\vz_\pi$ is the rendered image's latent.

Prior works have highlighted the drawbacks of employing a uniform timestep schedule in the optimization process of SDS \cite{graikos2022diffusion,yi2024diffusion,huang2023dreamtime}. 
The optimization process of SDS presents a dilemma: a large noise scale is necessary in the early stages to tackle the out-of-domain issue of the rendered images. 
However, this large noise scale also corrupts the information of the original images, making the optimization process unstable. 
Therefore, it is reasonable to adopt an annealing timestep schedule so that coarse-grained structures are generated at the early stage with a large noise level, while intricate details are refined at a later stage with a lower noise level.

To further unveil the connections between SDS optimization and DDCM sampling, we make a reasonable assumption that the timestep $t$ should follow an annealing schedule, as suggested by extensive prior research.
We can write the pseudo-ground-truth as a function of the rendered image's latent $\vz_\pi$, timestep $t$, condition $y$ and added noise $\vepsilon$:
\begin{equation}
    \begin{aligned}
        \zpred{t} & \coloneq \vf_\phi(\vz_\pi, \vepsilon, t, y)
        \\
        &= \frac{(\vz_t - \std{t} \vepsilon_\phi(\vz_t,t,y))}{\scale{t}}
        \\
        &= \frac{\left(\scale{t} \vz_\pi + \std{t} \vepsilon - \std{t} \vepsilon_\phi(\vz_t,t,y)\right)}{\scale{t}}
        \\
        &= \vz_\pi + \frac{\std{t}}{\scale{t}} \left(\vepsilon - \vepsilon_\phi(\scale{t} \vz_\pi + \std{t} \vepsilon,t,y)\right) \label{eqn:pseudo-gt} \; .
    \end{aligned}
\end{equation}
Note that \refe{sample-ddcm} in DDCM describes the dynamics of $\vz_t$. We reparameterize it to model the dynamics of $\zpred{t}$ instead, formulated as:
\begin{equation}
    \begin{aligned}
    \zpred{t} &= \frac{\vz_{t} - \sqrt{1-\alpha_{t}} \vepsilon_\phi}{\sqrt{\alpha_{t}}}
    \\
    &= \frac{\sqrt{\alpha_{t}} \zpred{t+1} + \sqrt{1 - \alpha_{t}} \vepsilon_{t+1} - \sqrt{1 - \alpha_{t}} \vepsilon_\phi}{\sqrt{\alpha_{t}}}
    \\
    &= \zpred{t+1} + \frac{\sqrt{1 - \alpha_{t}}}{\sqrt{\alpha_{t}}} \left( \vepsilon_{t+1} - \vepsilon_\phi \right) \label{eqn:sample-ddcm-pred} \; .
    \end{aligned}
\end{equation}
We observe that the dynamics described by \refe{sample-ddcm-pred} closely resembles the optimization of SDS \refe{pseudo-gt}. 
With the assumption that SDS following an annealing timestep schedule, the SDS optimization is equivalent to an \textit{iterative reconstruction} procedure: \textbf{1)} sample next-step pseudo-ground-truth using noisy latents produced from previous rendered images with DDCM; \textbf{2)}  optimize a reconstruction loss to match the view with pseudo-ground-truth. This finding allows us to consider the design of 3D editing methods from a more general \textit{iterative reconstruction} perspective.
We expand this formulation to include variants of SDS, summarizing their pseudo-ground-truth parameterizations in \reft{f-variants}.

\begin{table}[tb]
  \caption{
  Denote $\gamma_t = \std{t} / \scale{t}$, we compare different pseudo-ground-truths of SDS and its variants.}
  \label{table:f-variants}
  \centering
  \resizebox{\linewidth}{!}{
  \begin{tabular}{ll}
    \toprule
    Methods     &   Pseudo-ground-truths $\vf$     \\
    \midrule
    SDS / SJC \cite{poole2023dreamfusion,wang2023score} & $\vz_\pi + \gamma_t\left(\vepsilon - \vepsilon_\phi(\vz_t, t, y)\right)$ \\
    VSD \cite{wang2023prolificdreamer} & $\vz_\pi + \gamma_t\left(\vepsilon_\varphi(\vz_t, t, y) - \vepsilon_\phi(\vz_t, t, y)\right)$  \\
    DDS \cite{hertz2023delta} & $\vz_\pi + \gamma_t\left(\vepsilon_\phi(\vz_t', t, y') - \vepsilon_\phi(\vz_t, t, y)\right)$    \\
    ISM \cite{liang2024luciddreamer}     & $\vz_\pi + \gamma_t\left(\vepsilon_\phi(\vz_s, s, \varnothing) - \vepsilon_\phi(\vz_t, t, y)\right)$ \\
    NFSD \cite{katzir2024noisefree}     & $\vz_\pi + \gamma_t\left(\vepsilon_\phi(\vz_t, t, y_{neg}) -  \vepsilon_\phi(\vz_t, t, \varnothing) - s\vdelta_\phi^C \right)$ \\
    \bottomrule
  \end{tabular}
  }
\end{table}

\vspace{-8pt}

\subsection{Trajectory-Anchored Scheme for Progressive 3D Gaussian Editing}

Our previous analysis in \refs{analysis} bridges the gap between \textit{optimization-based} and \textit{reconstruction-based} 3D Gaussian editing methods, demonstrating that optimization-based approaches are special cases of reconstruction-based methods.
Without loss of generality, we consider the design of 3D editing methods from a broader \textit{reconstruction-based} perspective. The key questions for designing improved 3D Gaussian editing methods become: \textit{
(1) which appropriate reconstruction pseudo-ground truths to use,
and (2) how to schedule the reconstruction process of 3D Gaussians in a progressive manner for 3D editing.
}

With this in mind, we propose a trajectory-anchored progressive 3D Gaussian editing scheme, as shown in \reff{framework}.
This approach more tightly couples the image editing process with the rendering process, enabling 3D rendering to actively rectify the inconsistencies in the progressive 2D multi-view edits.
Specifically, we use images generated by a modified DDCM editing trajectory, similar to InfEdit \cite{xu2024inversionfree}, as suitable pseudo-ground-truths for 3D Gaussian editing. 
Consider the DDCM pseudo-inversion process in InfEdit \cite{xu2024inversionfree}:
\begin{equation}
    \ztgt{0} \coloneq \frac{\ztgt{t} - \std{t} (\epsilontgt - \epsilonsrc + \epsilon)}{\scale{t}}
\end{equation}
where $\epsilonsrc$ and $\epsilontgt$ are the noise prediction from the source branch and target branch, respectively.
Let 
\begin{equation}
    \begin{aligned}
    &\zsrccons{0} \coloneq \frac{\zsrc{t_n} - \std{t_n} \epsilonsrc}{\scale{t_n}}
    \\
    &\ztgtcons{0} \coloneq \frac{\ztgt{t_n} - \std{t_n} \epsilontgt}{\scale{t_n}}
    \end{aligned} \; ,
\end{equation}
the pseudo-ground-truth $\ztgt{0}$ can be reparameterized as
\begin{equation}
    \begin{aligned}
        \ztgt{0} = \ztgtcons{0} + \kappa \left(\zsrc{0} - \zsrccons{0} \right)
    \end{aligned} \; .
    \label{eqn:ddcm-pseudo-gt}
\end{equation}
Here, $\kappa$ denotes a DDCM adjustment coefficient. When $\kappa = 1$, the pseudo-ground-truth trajectory coincides with the editing trajectory of InfEdit. 
We observe that the second term in \refe{ddcm-pseudo-gt} injects information from the original image into $\ztgtcons{0}$ during the editing process.
Consequently, a large $\kappa$ may lead to an ``overshooting" effect, resulting in over-saturated color and over-sharpened edges, analogous to image sharpening. 
In our experiments, we evaluate the impact of various values of $\kappa$ on 2D view editing and 3D scene editing and select the optimal value for our method.

The detailed sampling process for generating pseudo-ground-truths is described in \refa{tas}.
These pseudo-ground-truths enable a smooth transition from the source image to the target image (as illustrated in \reff{trajectory}), making them ideal for the progressive 2D editing and 3D updating scheme.
Progressive edits on 2D views can be promptly applied to 3D Gaussians.
Conversely, minor view inconsistencies arising from 2D editing can be promptly rectified through 3D constrained rendering.
\begin{figure*}[t]
  \centering
  \includegraphics[width=\textwidth]{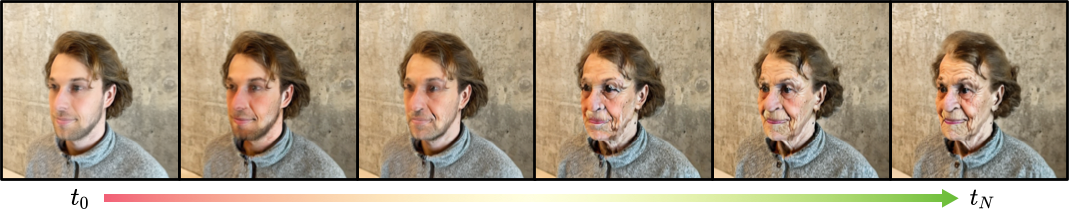}
  \vspace{-10pt}
  \caption{The editing trajectory exhibits a smooth and incremental transition from the original image to the final edited image.}
  \vspace{-10pt}
  \label{fig:trajectory}
\end{figure*}
In line with GaussianEditor \cite{chen2023gaussianeditor}, we adopt a combination of reconstruction $\mathcal{L}_1$ loss, perceptual loss $\mathcal{L}_\text{LPIPS}$, and anchor loss $\mathcal{L}_\text{anchor}$ as our loss function $\mathcal{L}$:
\begin{equation}
    \label{eqn:loss}
    \mathcal{L} = \mathcal{L}_1 + \lambda_\text{LPIPS} \cdot \mathcal{L}_\text{LPIPS} + \lambda_\text{anchor} \cdot \mathcal{L}_\text{anchor} \; ,
\end{equation}
where $\lambda_\text{LPIPS}$ and $\lambda_\text{anchor}$ respectively denote the weight coefficient of perceptual loss and anchor loss.

\begin{algorithm}[tb]
\caption{Trajectory-Anchored Scheme}
\label{alg:tas}
\textbf{Input:}
$\thetasrc$, 
$\left\{t_n\right\}_{n=1}^N$,
$\ysrc$, $\ytgt$,
$\eta$,
$K$,
$\kappa$

\begin{algorithmic}
\State {\bfseries initialize} $\thetat{t_n} \leftarrow \thetasrc$.

 \For{$n=1$ {\bfseries to} $N$}
    \State Sample $\vepsilon \sim \mathcal{N}(\bm{0}, \mI)$
    \State $\zpi \leftarrow \vg(\thetat{t_n})$ 
    \State $\ztgt{t_n} \leftarrow \scale{t_n} \zpi + \std{t_n} \vepsilon $ 
    \State $\zsrc{t_n} \leftarrow \scale{t_n} \zsrc{0} + \std{t_n} \vepsilon $ 
    \State $\zsrccons{0} \leftarrow \left(\zsrc{t_n} - \std{t_n} \epsilonsrc \right) / \scale{t_n}$
    \State $\ztgtcons{0} \leftarrow \left(\ztgt{t_n} - \std{t_n} \epsilontgt \right) / \scale{t_n}$
    \State $\ztgt{0} \leftarrow \ztgtcons{0} + \kappa \left( \zsrc{0} - \zsrccons{0} \right)$
     \For{$k=1$ {\bfseries to} $K$}
        \State $\zpi \leftarrow \vg(\thetat{t_n})$
        \State $\thetat{t_n} \leftarrow \thetat{t_n} - \eta \nabla_{\theta} \mathcal{L}(\zpi, \ztgt{0})$ 
     \EndFor

     \State $\thetat{t_{n-1}} \leftarrow \thetat{t_{n}}$

 \EndFor

\State {\bfseries return $\thetat{t_1}$} 
\end{algorithmic}
\end{algorithm}

\vspace{-11pt}

\subsection{View-Consistent Attention Control}
\label{sec:attn-control}
\vspace{-2pt}

Ensuring consistent edits across multiple views is crucial for effective optimization of 3D Gaussians, as inconsistent edits among different views may fail to consolidate in 3D scenes \cite{haque2023instructnerfnerf}. To address this, we proposed a dual-branch editing scheme that utilizes structural information from the original views as 3D view prior, imposing structural constraints on the spatial layouts of the target views. Furthermore, we employed KV propagation and KV reference mechanisms to enhance semantic consistencies across views.

We utilize a source branch and a target branch to achieve 3D consistent editing. The source branch takes in unedited original views, acting as a reference for the target branch, providing structural and semantic guidance for generating edited views.
Previous studies have noted that the structural layout of an image is formed in the early stages of diffusion and is closely linked to self-attention queries. To ensure that the edited views produced by the target branch maintain structural multi-view consistency, we opt to inject self-attention queries from the source branch into the target branch during the early stages of diffusion (mid-bottom part of \reff{framework}). Specifically, denote the attention calculation as:
\begin{equation}
\label{eqn:attn}
\text{Attn}(\mQ,\mK,\mV) = \softmax\left(\frac{\mQ {\mK}^\top}{\sqrt{d}}\right) \mV \,.
\end{equation}
The query injection is performed as:
\begin{equation}
\label{eqn:q-injection}
\text{QInj}(\Qsrc,\!\Qtgt,\!\Ktgt,\!\Vtgt,\!t)\!=\!
\left\{\!
\begin{array}{ll}
\!\text{Attn}(\Qsrc,\!\Ktgt,\!\Vtgt)&\!t\!\le\!t_{q}
\\
\!\text{Attn}(\Qtgt,\!\Ktgt,\!\Vtgt)&\!t\!>\!t_{q}
\end{array}
\right.,
\end{equation}
where $t_{q}$ is the threshold that determines the number of steps for which the injection applies.

Maintaining structural multi-view consistency alone is insufficient; semantic multi-view consistency is also crucial. To achieve this, we implemented KV propagation and KV reference mechanisms (bottom-left of \reff{framework}). 
Given a sequence of $N$ views $\{\mI^n\}_{n=1}^{N}$ resulting from consecutive camera motion, we divide the sequence into equal-partitioned context segments $\{\mC^i\}_{i=1}^{L}$.
Each context $\mC^i$ has a length of $N / L$ frames and contains a keyframe $\mI_1^i$.
Views within the same context exhibit minimal movement or view shift, their semantics are relatively consistent.
For views within the same context, we utilized an efficient KV propagation method, where the self-attention keys and values of the keyframe $\{ \mK_1^i,\mV_1^i \}$ of $i$-th context is directly propagated to the rest of the frames in the same context.
Denote the output of self-attention of $j$-th frame in context $i$ as $\mO^i_j$, we formulate this attention propagation as:
\begin{equation}
    \mO^i_j = \text{Attn} \left(\mQ_j^i, \mK_1^i, \mV_1^i \right) \; .
\end{equation}
Significant view shifts occur across different contexts, leading to noticeable semantic differences.
Applying KV propagation on keyframes across different contexts would accumulate significant distortion.
To mitigate this, we employ KV reference for keyframes across contexts. 
This involves inflating the self-attention keys and values across keyframes of different contexts and processing them jointly.
The self-attention keys and values from other keyframes serve as a mutual reference for the semantic information of the edited entity.
For self-attention keys $\{\mK^i\}_{i=1}^{l}$ and values $\{\mV^i\}_{i=1}^l$ from $l$ different keyframes, we concatenate them respectively to form cross-context KV pairs.
The operation is formulated as:
\begin{equation}
\label{eqn:kv-keyframe}
\mO_1^i = \text{Attn} \left(\mQ_1^i, \left[\mK^1, \dots, \mK^l\right], \left[\mV^1, \dots, \mV^l\right]\right) \; .
\end{equation}

To maintain fidelity with the original images in the non-edited areas, we apply a cross-attention control (bottom-right of \reff{framework}) and local blending strategy similar to Prompt-to-Prompt \cite{hertz2022prompttoprompt} and InfEdit \cite{xu2024inversionfree}. 
Cross-attention is done by replacing the attention maps corresponding to non-edited tokens in the target branch with those from the source branch.
We use the semantic mask $\mM$ obtained from semantic tracing \cite{chen2023gaussianeditor} to blend noisy target latent with source latent, formulated as:
\begin{equation}
\label{eqn:local-blending}
    \ztgt{t} = \mM \odot \ztgt{t} + (1 - \mM) \odot \zsrc{t} \; ,
\end{equation}
where $\odot$ is the Hadamard product.
Consequently, the 3D consistent view edits consolidate onto the 3D Gaussians in a rectified and progressive manner.

\section{Experiments}
\label{sec:experiments}

\vspace{-5pt}

\begin{table*}[tb]
\centering
\caption{Quantitative Results. Comparative analysis on the reconstruction-based methods regarding the coherence between 2D view edits and 3D scene modifications.}
\label{table:ch5:comparsion_consistency}
\resizebox{\linewidth}{!}{
\begin{tabular}{lcccccc} 
\toprule
\multirow{2}{*}{Methods} & \multirow{2}{*}{PSNR $\uparrow$} & \multirow{2}{*}{SRE $\uparrow$} & \multirow{2}{*}{RMSE ($\times 10^{-5}$) $\downarrow$} & \multicolumn{3}{c}{CLIP Image-to-Image Similarity $\uparrow$}               \\ 
\cline{5-7}
\rule{0pt}{9pt}  
                  &                  &                &             & ViT-B/16 & ViT-B/32 & ViT-L/14  \\ 
\midrule
IN2N \cite{haque2023instructnerfnerf}              & 87.860 \tiny{$\pm$1.050}           & 29.828 \tiny{$\pm$0.771}          & 4.438 \tiny{$\pm$0.509}          & 0.888 \tiny{$\pm$0.036}               & 0.896 \tiny{$\pm$0.029}         & 0.866 \tiny{$\pm$0.033}                \\
VICA \cite{dong2023vicanerf}             & 88.846 \tiny{$\pm$0.676}           & 30.369 \tiny{$\pm$0.380}          & 3.892 \tiny{$\pm$0.246}         & 0.880 \tiny{$\pm$0.024}              & 0.867 \tiny{$\pm$0.023}         & 0.858 \tiny{$\pm$0.021}                \\
GSEditor \cite{chen2023gaussianeditor}         & 96.040 \tiny{$\pm$1.755}          & 34.487 \tiny{$\pm$0.991}         & 1.649 \tiny{$\pm$0.357}         & 0.939 \tiny{$\pm$0.016}              & 0.936 \tiny{$\pm$0.017}         & 0.914 \tiny{$\pm$0.026}                \\ 
\midrule
TrAME (w/o TAS \& w/o VCAC) & 87.214 \tiny{$\pm$1.444}          & 30.575 \tiny{$\pm$0.713}         & 4.438 \tiny{$\pm$0.720}         & 0.669 \tiny{$\pm$0.053}              & 0.675 \tiny{$\pm$0.066}      & 0.601 \tiny{$\pm$0.053}                \\
TrAME (w/o TAS \& w. VCAC)           & 90.101 \tiny{$\pm$0.988}          & 32.283 \tiny{$\pm$0.581}         & 3.169 \tiny{$\pm$0.350}         & 0.801 \tiny{$\pm$0.061}              & 0.804 \tiny{$\pm$0.058}        & 0.754 \tiny{$\pm$0.061}                \\
TrAME (w. TAS \& w. VCAC)              & \textbf{102.449 \tiny{$\pm$0.499}} & \textbf{38.322 \tiny{$\pm$0.247}} & \textbf{0.758 \tiny{$\pm$0.044}}  & \textbf{0.962 \tiny{$\pm$0.005}}      & \textbf{0.978 \tiny{$\pm$0.003}} & \textbf{0.937\tiny{$\pm$0.011}}       \\
\bottomrule
\end{tabular}
}
\end{table*}

In this section, we provide a comprehensive overview of our experimental setup, including the testing data, baseline methods, and implementation details.
We evaluate the effectiveness of our method using both quantitative and qualitative results. 
Additionally, we conduct ablation studies on each proposed component to analyze their individual contributions to overall performance.
In the final part, we further analyze the limitations and broader impact of our work.
\vspace{-5pt}

\subsection{Implementation Details}
\label{sec:implementation}


We implement our method based on a modified version of GaussianEditor \cite{chen2023gaussianeditor}.
Each editing typically takes about 5-15 minutes on average, depending on the complexity of the scene.
For scenes involving partial editing, we adhere to the semantic tracing approach outlined in GaussianEditor.
The context length of the VCAC module is set to 4.
A view angle difference threshold of 25 degrees is set to determine if the displacement between two views is too large, thereby deciding whether to use the KV propagation or KV reference approach in the VCAC module.
We evaluate our method on Instruct-N2N \cite{haque2023instructnerfnerf} and MipNeRF-360 \cite{barron2022mip} dataset.
The experiments are conducted using a single Nvidia A100 80G GPU, with VRAM consumption around 30GB.

\vspace{-5pt}

\subsection{Baseline Methods}


We conduct a comparative analysis of our method against five existing 3D scene editing approaches. 
Specifically, Instruct-NeRF2NeRF (IN2N) \cite{haque2023instructnerfnerf}, ViCA-NeRF (VICA) \cite{dong2023vicanerf}, and Posterior Distillation Sampling (PDS) \cite{koo2023posterior} employ NeRF as a 3D representation, while GaussianEditor (GSEditor) \cite{chen2023gaussianeditor} and Delta Denoising Score (DDS) \cite{hertz2023delta} utilize 3DGS for this purpose.
Furthermore, PDS and DDS serve as optimization-based 3D editing baselines, whereas the remaining methods are reconstruction-based.
Our quantitative evaluation is based on the CLIP Score and CLIP Directional Score \cite{brooks2023instructpixpix}, assessing the semantic coherence between the edited 3DGS and the given prompt.
In addition, for the reconstruction-based methods, we measure image-to-image similarity between the edited and rendered views of the scenes, to evaluate the consistency between 2D view edits and 3D scene modifications.

\vspace{-5pt}

\begin{table}[tb]
\centering
\caption{Quantitative Results: CLIP Directional Score against NeRF-based methods (top) and 3DGS-based methods (middle).}
\label{table:ch5:comparison}
\resizebox{\columnwidth}{!}{
\begin{tabular}{lccc} 
\toprule
\multirow{2}{*}{Methods}
& \multicolumn{3}{c}{CLIP Directional Score} 
\\ 
\cline{2-4}
\rule{0pt}{9pt}  
& ViT-B/16 $\uparrow$ 
& ViT-B/32 $\uparrow$ 
& ViT-L/14 $\uparrow$ 
\\ 
\midrule
IN2N \cite{haque2023instructnerfnerf}       & 0.148 \tiny{$\pm$0.068} & 0.149 \tiny{$\pm$0.060} & 0.127 \tiny{$\pm$0.064} \\
VICA \cite{dong2023vicanerf}     & 0.157 \tiny{$\pm$0.052} & 0.151 \tiny{$\pm$0.044} & 0.145 \tiny{$\pm$0.044} \\
PDS  \cite{koo2023posterior}      & 0.163 \tiny{$\pm$0.080} & 0.160 \tiny{$\pm$0.075} & 0.155 \tiny{$\pm$0.066} \\ 
\midrule
DDS  \cite{hertz2023delta}      & 0.121 \tiny{$\pm$0.074} & 0.134 \tiny{$\pm$0.075} & 0.111 \tiny{$\pm$0.071} \\
GSEditor \cite{chen2023gaussianeditor}  & 0.166 \tiny{$\pm$0.056} & 0.162 \tiny{$\pm$0.055} & 0.161 \tiny{$\pm$0.054} \\ 
\midrule
TrAME (Ours)       & \textbf{0.199 \tiny{$\pm$0.054}} & \textbf{0.201 \tiny{$\pm$0.055}} & \textbf{0.186 \tiny{$\pm$0.049}} \\
\bottomrule
\end{tabular}
}
\end{table}

\begin{figure*}[tb]
  \centering
  \includegraphics[width=\textwidth]{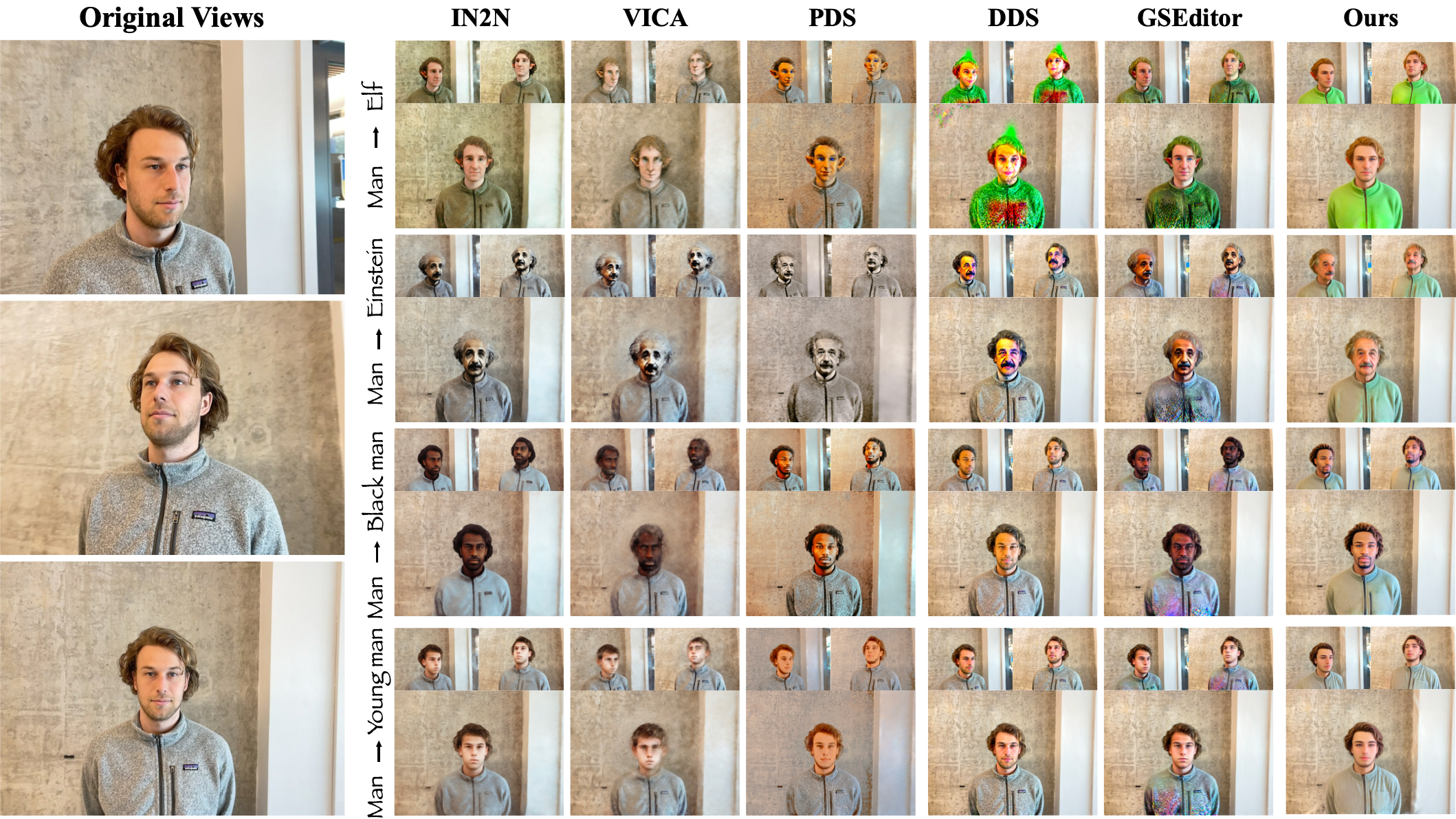}
  \vspace{-10pt}
  \caption{Qualitative comparison. A comparative analysis of experimental results for single-scene editing across State-of-the-Art methods and ours. Please zoom in for more geometry and textural details.
  }
  \label{fig:ch5:exp:comparsion:man}
  \vspace{-10pt}
\end{figure*}

\vspace{-5pt}

\begin{figure}[tb]
  \centering
  \includegraphics[width=\linewidth]{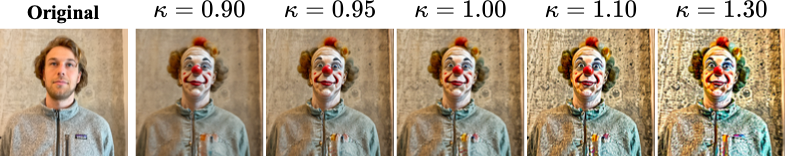}
  \vspace{-10pt}
  \caption{Qualitative comparison. A comparative analysis of different design choices of DDCM coefficient $\kappa$ on 2D view editing.}
  \label{fig:ddcm-kappa-2d}
\end{figure}

\begin{figure}[tb]
  \centering
  \includegraphics[width=\linewidth]{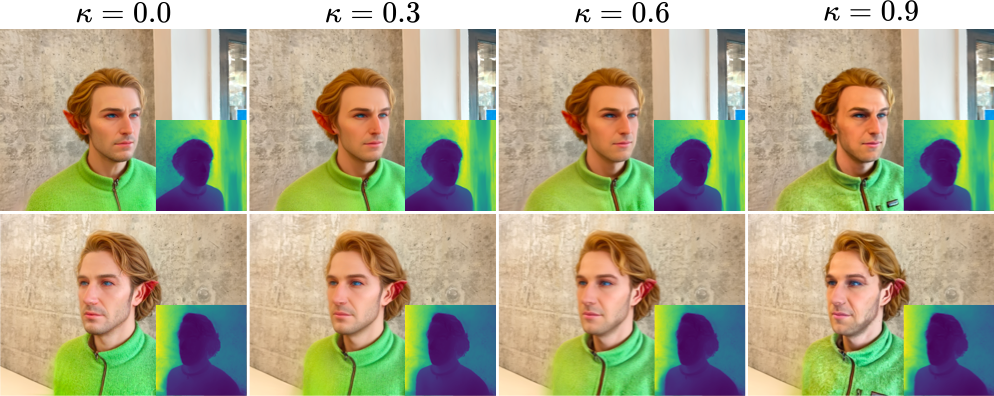}
  \vspace{-10pt}
  \caption{Qualitative results under different hyper-parameter settings. Rendered RGB and depth representations illustrating the implications of different DDCM coefficient $\kappa$ on 3DGS editing.}
  \label{fig:ddcm-kappa-3d}
\end{figure}

\subsection{Quantitative Results}


In our quantitative analysis, we employ the CLIP Score and CLIP Directional Score \cite{brooks2023instructpixpix} to assess the alignment of the edited 3D models with the target text prompts on the rendered views of the edited 3DGS.
These results indicate that our 3D editing method is effective at consolidating user edits into 3D scenes.
The quantitative evaluation results, presented in \reft{ch5:comparison}, compare our method against baseline methods across various test scenes. 
Our method excels in CLIP Score, CLIP Directional Score, confirming its superior editing outcomes.

Additionally, for reconstruction-based methods, we assess the coherence between 2D view edits and the rendered views of the modified 3D scenes by calculating image-to-image similarity. This is achieved by measuring peak signal-to-noise ratio (PSNR), signal-to-reconstruction error ratio (SRE), root mean square error (RMSE), and CLIP image-to-image similarity using three different backbones: ViT-B/16, ViT-B/32, and ViT-L/14.
The results depicted in \reft{ch5:comparsion_consistency} demonstrate that our method achieves greater consistency between 2D view edits and 3D scene modifications, underscoring the effectiveness of our approach.

\vspace{-5pt}

\subsection{Qualitative Results}


As shown in \reff{ch5:exp:comparsion:man} and \reff{ch5:exp:comparsion:scene}, our qualitative analysis showcases that our method ensures multi-view consistency and fine-grained details in 3D scene editing.
We display outcomes from various perspectives in different scenes, underscoring the effectiveness of our method.
The qualitative results reveal that IN2N, VICA and PDS struggle to produce accurate and localized edits, while DDS faces challenges in achieving detailed and faithful editing.
VICA, in particular, often generates blurry results due to the errors of depth estimation. 
PDS occasionally fails to produce correct scene colors, as depicted in the ``Hulk" case, ``Golden bear" case and the ``Park" case in \reff{ch5:exp:comparsion:scene}.
Moreover, the inconsistent view edits of GSEditor often result in unstable optimization, leading to visible artifacts, such as the noisy colorization of the man's jacket in \reff{ch5:exp:comparsion:man} and the ``autumn park" case in \reff{ch5:exp:comparsion:scene}.
These shortcomings of the compared methods become particularly apparent when trying to capture the intricate patterns and fine details inherent to such materials, highlighting a critical area where our method excels in producing view-consistent and detailed 3D scene edits.

\begin{figure*}[htb!]
  \centering
  \includegraphics[width=0.95\textwidth]{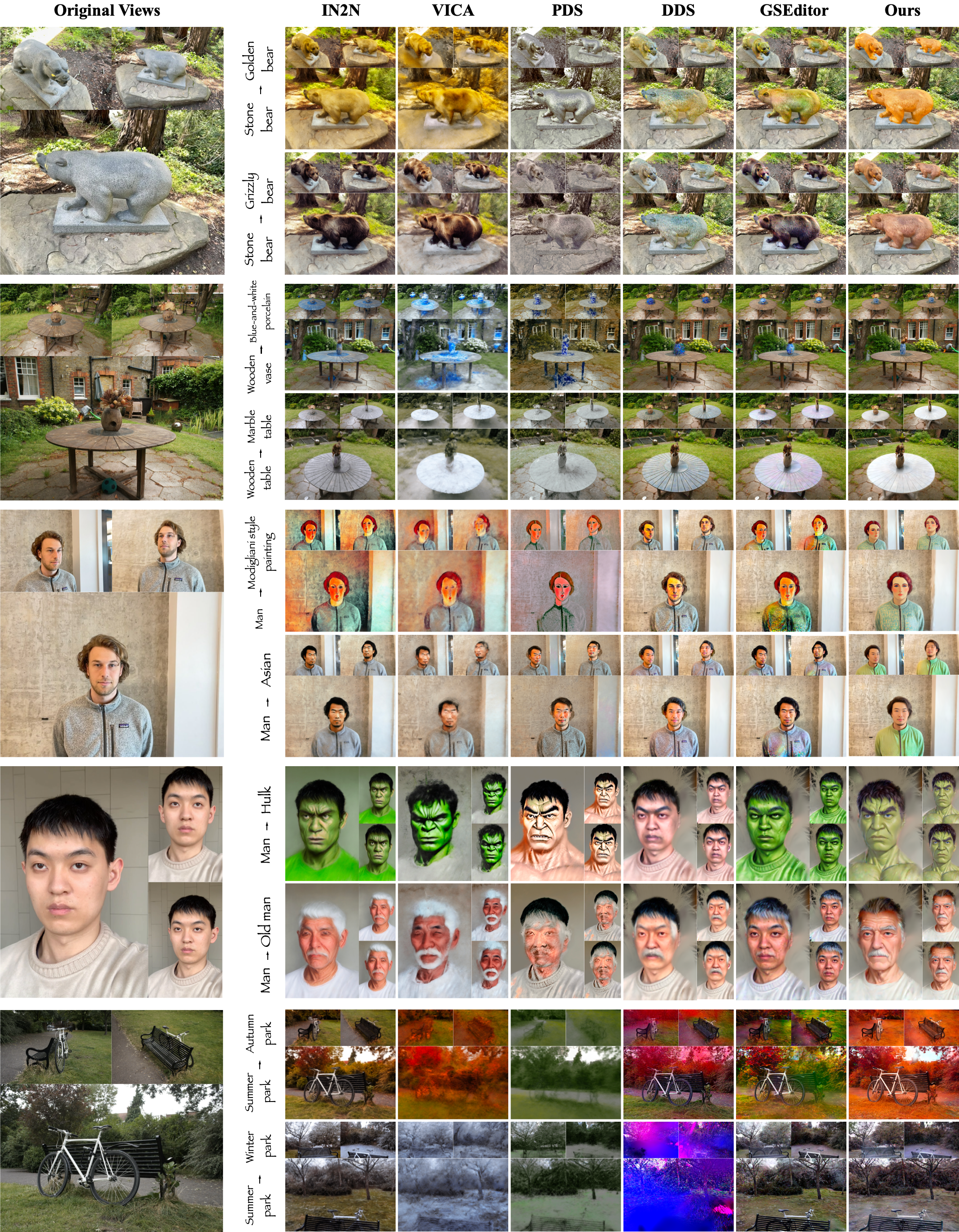}
  \caption{
  Qualitative comparison. A comparative analysis of experimental results for various scenes encompassing both partial and global editing. Please zoom in for more geometry and textural details.}
  \label{fig:ch5:exp:comparsion:scene}
\end{figure*}

\vspace{-5pt}

\subsection{Ablation Studies}

\begin{figure}[t]
  \centering
  \includegraphics[width=\linewidth]{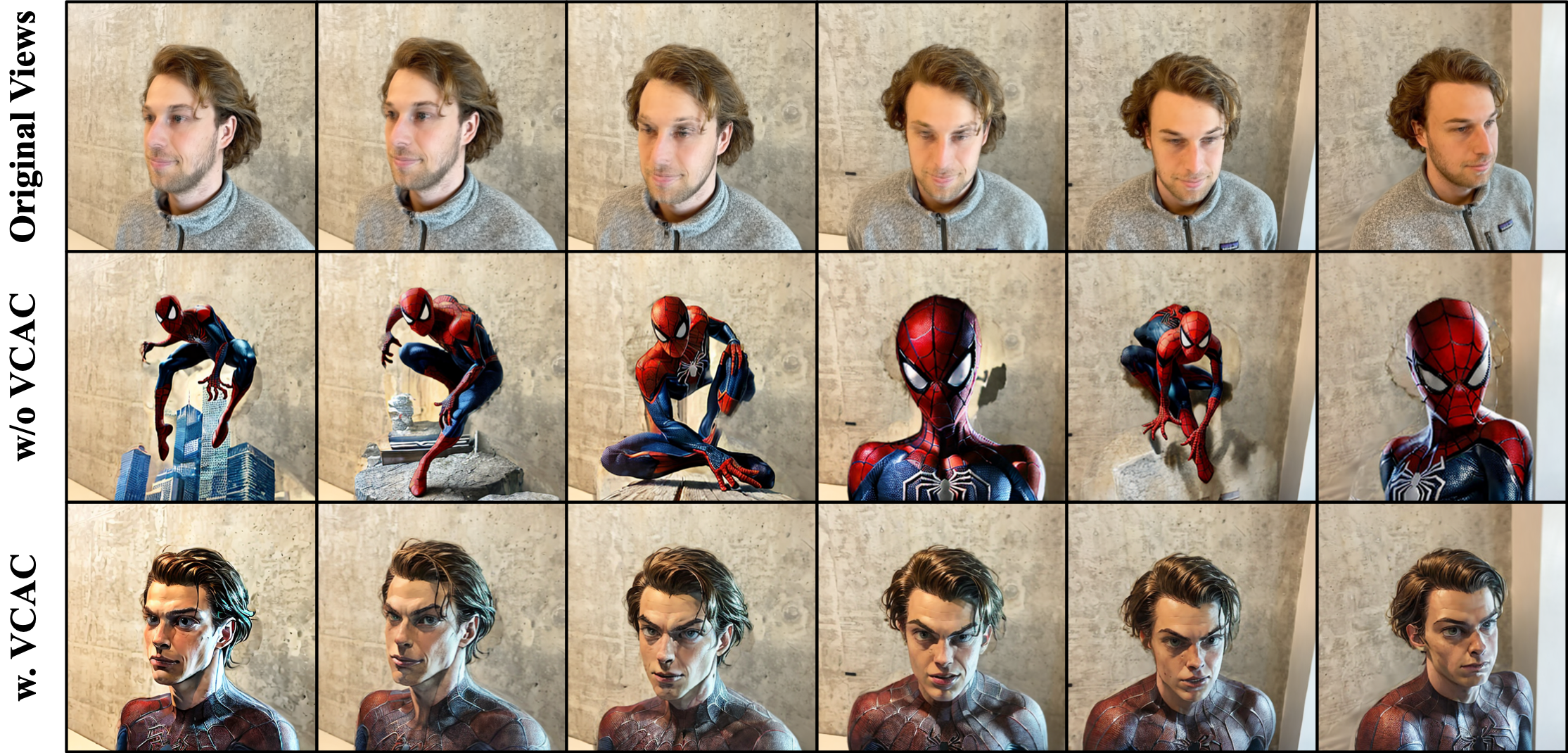}
  \vspace{-10pt}
  \caption{Ablation study on the VCAC module for 2D view editing. The results display edited views with (w.) and without the VCAC module (w/o).}
  \label{fig:2d-ablation}
  \vspace{-5pt}
\end{figure}

\begin{figure}[t]
  \centering
  \includegraphics[width=\linewidth]{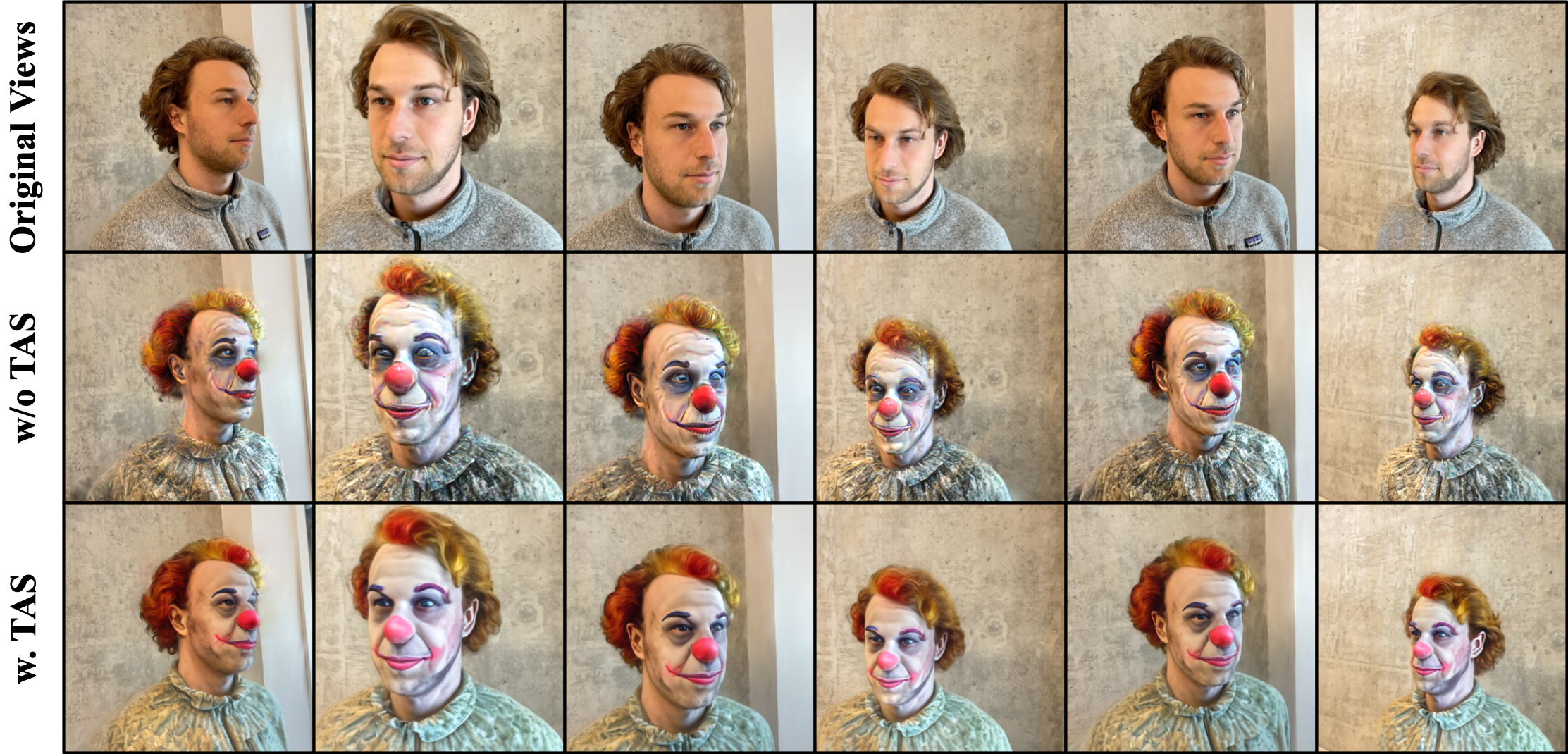}
  \vspace{-10pt}
  \caption{Ablation study on TAS for 3DGS editing. Comparative results are shown with (w. TAS) and without the adoption of the TAS (w/o TAS).}
  \vspace{-8pt}
  \label{fig:3d-ablation}
\end{figure}


\textbf{Study on VCAC.} The qualitative results presented in \reff{2d-ablation} demonstrate that the VCAC module effectively enhances cross-view structural and semantic consistency. 
The views modified by our VCAC module align more accurately with the original views in terms of structural elements like poses and facing directions, and they also demonstrate better semantic coherence, exhibiting similar appearances.
In particular, the outputs generated by our VCAC module maintain a uniform appearance across various views, with the facing direction of Spiderman in the figures aligning cohesively with the pose of the man in the original views. Additionally, the quantitative results in \reft{ch5:comparsion_consistency} indicate that the VCAC module significantly reduces inconsistencies between 2D edited views and 3D scene modifications by producing consistent 2D view edits, thereby further validating its effectiveness in preserving coherence across different views.

\textbf{Study on TAS.} The ablated results obtained without TAS were derived by applying 3D editing solely based on the final obtained edited views, without feedback from the 3D Gaussians.
The qualitative results depicted in \reff{3d-ablation} show that TAS produces more naturalistic and consistent appearances, while the ablated results exhibit undesirable blurry artifacts and inconsistencies across views. Specifically, the hair color around the clown's forehead in the ablated results varies across views, and the lips of the clown show ghosting shadows.
Moreover, the quantitative results in \reft{ch5:comparsion_consistency} confirm that our proposed TAS effectively mitigates accumulated errors during the 2D view editing process, thereby resulting in higher consistency with the rendered views of the modified 3D scenes.

\textbf{Study on DDCM coefficient $\kappa$.} We conducted a qualitative analysis of the effects of varying $\kappa$ values on both 2D view and 3D scene editing. The results are presented in \reff{ddcm-kappa-2d} and \reff{ddcm-kappa-3d}. As shown in \reff{ddcm-kappa-2d}, selecting a $\kappa$ value greater or equal to 1.0 often results in over-saturated colors and excessively sharpened edges, thereby reducing the fidelity of the edited image. The optimal value of $\kappa$ was determined to be around 0.95, which yields the highest visual quality. Additionally, we observed that using a larger $\kappa$ in later stages of editing may hinder the ability to properly edit geometry. For instance, in \reff{ddcm-kappa-3d}, the elf's right ear does not consolidate effectively with $\kappa=0.9$. Hence, a smaller $\kappa$ should be applied during the later stages of editing.

\textbf{Study on query injection threshold $t_q$.} We quantitatively analyze the effect of varying self-attention query injection thresholds. The results, presented in \reft{ablation-tq}, reveal that performance is optimal around $t_q = 0.5$ and deteriorates as $t_q$ increases. This decline is attributed to the over-constraint imposed by query injection. Specifically, query injection tends to impose excessive constraints on the edited images, resulting in reduced modifications to the edited views, which eventually leads to diminished overall editing effectiveness.

\vspace{-5pt}

\subsection{Limitations and Broader impact}
\textbf{Limitations.}
Our limitations lie in performing complex, view-consistent edits involving significant object deformations.
As illustrated in \reff{failure-cases}, both our approach and existing baselines encounter difficulties with scene edits that necessitate extensive deformation or displacement. 
This limitation is primarily due to the lack of 3D priors in 2D image editing models and the insufficient integration of 3D geometry required to achieve the desired edits. 
In future work, we aim to leverage the explicit geometry of the 3D Gaussian Splatting to incorporate multi-level hierarchical editing guidance, facilitating the execution of complex edits in a coarse-to-fine manner.

\textbf{Broader impact.} 
Our work has the potential to significantly advance the creation of immersive virtual environments, thereby enhancing applications in gaming, virtual reality, and film production.
However, the capability to seamlessly edit and manipulate 3D scenes also carries the risk of being exploited to create misleading or harmful visual content. 
This is particularly concerning in contexts where authenticity is crucial, such as in news media or scientific visualization.

\begin{table}[tb]
\scriptsize
\centering
\caption{Quantitative experiment for evaluating the CLIP Directional Score on rendered views of edited 3D scenes with varying values of $t_q$.}
\label{table:ablation-tq}
\resizebox{\columnwidth}{!}{
\begin{tabular}{lccc}
\toprule
& \multicolumn{3}{c}{CLIP Directional Score} 
\\ 
\cline{2-4}
\rule{0pt}{9pt}  
& ViT-B/16 $\uparrow$ 
& ViT-B/32 $\uparrow$ 
& ViT-L/14 $\uparrow$ 
\\ 
\midrule
$t_q = 0.4$ & 0.222\tiny{$\pm$0.049} & \textbf{0.191\tiny{$\pm$0.065}} & 0.195\tiny{$\pm$0.053} \\
$t_q = 0.5$ & \textbf{0.224\tiny{$\pm$0.051}} & 0.188\tiny{$\pm$0.067} & \textbf{0.195\tiny{$\pm$0.052}} \\
$t_q = 0.6$ & 0.215\tiny{$\pm$0.054} & 0.182\tiny{$\pm$0.072} & 0.185\tiny{$\pm$0.059} \\
$t_q = 0.7$ & 0.203\tiny{$\pm$0.063} & 0.180\tiny{$\pm$0.080} & 0.171\tiny{$\pm$0.078} \\
$t_q = 0.8$ & 0.191\tiny{$\pm$0.069} & 0.166\tiny{$\pm$0.083} & 0.162\tiny{$\pm$0.081} \\
$t_q = 0.9$ & 0.173\tiny{$\pm$0.084} & 0.149\tiny{$\pm$0.089} & 0.145\tiny{$\pm$0.086} \\
\bottomrule
\end{tabular}
}
\end{table}


\begin{figure}[t]
  \centering
  \includegraphics[width=\linewidth]{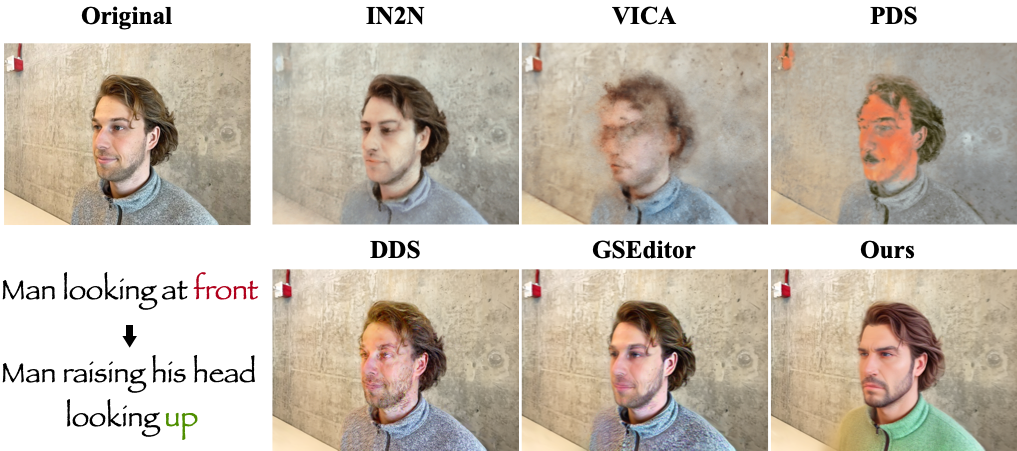}
  \vspace{-10pt}
  \caption{Failure cases. Our methods and existing baselines fails to perform complex edits that require large object deformations.}
  \vspace{-8pt}
  \label{fig:failure-cases}
\end{figure}

\section{Conclusion}
\label{sec:conclusion}


Our research introduces TrAME, a progressive 3DGS editing framework that effectively improves multi-view consistency via Trajectory Anchored Scheme (TAS) and View-Consistent Attention Control (VCAC) module. 
Our work bridges the gap between optimization-based and reconstruction-based editing methods, offering a unified perspective for selecting better design choices in 3D scene editing methods.

\vspace{-5pt}

\bibliographystyle{IEEEtran}
\bibliography{references}

\end{document}